\pdfoutput=1
\documentclass[11pt]{article} 
\usepackage{emnlp2023}

\usepackage{times}
\usepackage{latexsym}
\usepackage{tabularx}
\usepackage{multirow}
\usepackage{booktabs} 
\usepackage{flushend} 
\usepackage{balance} 
\usepackage{inconsolata}
\usepackage{graphicx}
\usepackage[utf8]{inputenc} 
\usepackage[T1]{fontenc}    
\usepackage{hyperref}       
\usepackage{url}            
\usepackage{booktabs}       
\usepackage{amsfonts}       
\usepackage{nicefrac}       
\usepackage{microtype}      
\usepackage{xcolor}         
\usepackage{multicol}
\usepackage{transparent}
\usepackage{array}
\usepackage{bm}
\usepackage{perpage} 
\MakePerPage{footnote} 
\usepackage{tabularx}
\usepackage{floatrow, makecell}
\usepackage{wrapfig,lipsum,booktabs}
\usepackage{pifont}
\usepackage{amssymb}
\usepackage{amsmath}
\usepackage{paralist}
\usepackage{caption}
\usepackage{subcaption}

\definecolor{midnightgreen}{rgb}{0.0, 0.29, 0.33}
\definecolor{lightgreen}{rgb}{0.0, 0.4, 0.20}
\definecolor{darkpink}{rgb}{0.91, 0.33, 0.5}
\definecolor{darkmagenta}{RGB}{139, 0, 139}
\definecolor{darkgreen}{rgb}{0,0.6,0}

\title{Rethinking Dense Retrieval's Few-Shot Ability}

\author{
\textbf{Si Sun}$^{\spadesuit}$\thanks{\text{   } Equal contributions.}, \textbf{Yida Lu}$^{\heartsuit*}$, \textbf{Shi Yu}$^{\heartsuit}$, \textbf{Xiangyang Li}$^{\clubsuit}$, \\
\textbf{Zhonghua Li}$^{\clubsuit}$, \textbf{Zhao Cao}$^{\clubsuit}$, \textbf{Zhiyuan Liu}$^{\heartsuit}$, \textbf{Deiming Ye}$^{\heartsuit}$, \textbf{Jie Bao}$^{\spadesuit}$ \\
\text{}$^{\spadesuit}$ \text{Dept. of Electron. Eng., Tsinghua University} \quad \text{}$^{\clubsuit}$ \text{Huawei Technologies Co., Ltd.} \\
\text{}$^{\heartsuit}$ \text{Dept. of Comp. Sci. \& Tech., Institute for AI, Tsinghua University} \\
\texttt{\{s-sun17, lu-yd20, yus21, ydm18\}@mails.tsinghua.edu.cn} \\
\texttt{\{lixiangyang21, lizhonghua3, caozhao1\}@huawei.com}; \texttt{\{liuzy, bao\}@tsinghua.edu.cn}
}

\begin{document}

\maketitle

\begin{abstract}
Few-shot dense retrieval (DR) aims to effectively generalize to novel search scenarios by learning a few samples. Despite its importance, there is little study on specialized datasets and standardized evaluation protocols. As a result, current methods often resort to random sampling from supervised datasets to create ``few-data'' setups and employ inconsistent training strategies during evaluations, which poses a challenge in accurately comparing recent progress. In this paper, we propose a customized FewDR dataset and a unified evaluation benchmark. Specifically, FewDR employs class-wise sampling to establish a standardized ``few-shot'' setting with finely-defined classes, reducing variability in multiple sampling rounds. Moreover, the dataset is disjointed into base and novel classes, allowing DR models to be continuously trained on ample data from base classes and a few samples in novel classes. This benchmark eliminates the risk of novel class leakage, providing a reliable estimation of the DR model's few-shot ability. Our extensive empirical results reveal that current state-of-the-art DR models still face challenges in the standard few-shot scene. Our code and data will be open-sourced at ~\url{https://github.com/OpenMatch/ANCE-Tele}.
\end{abstract}

\section{Introduction}

\begin{figure}[t]
\centering
\hspace{-0.58cm}
\includegraphics[height=3.9cm]{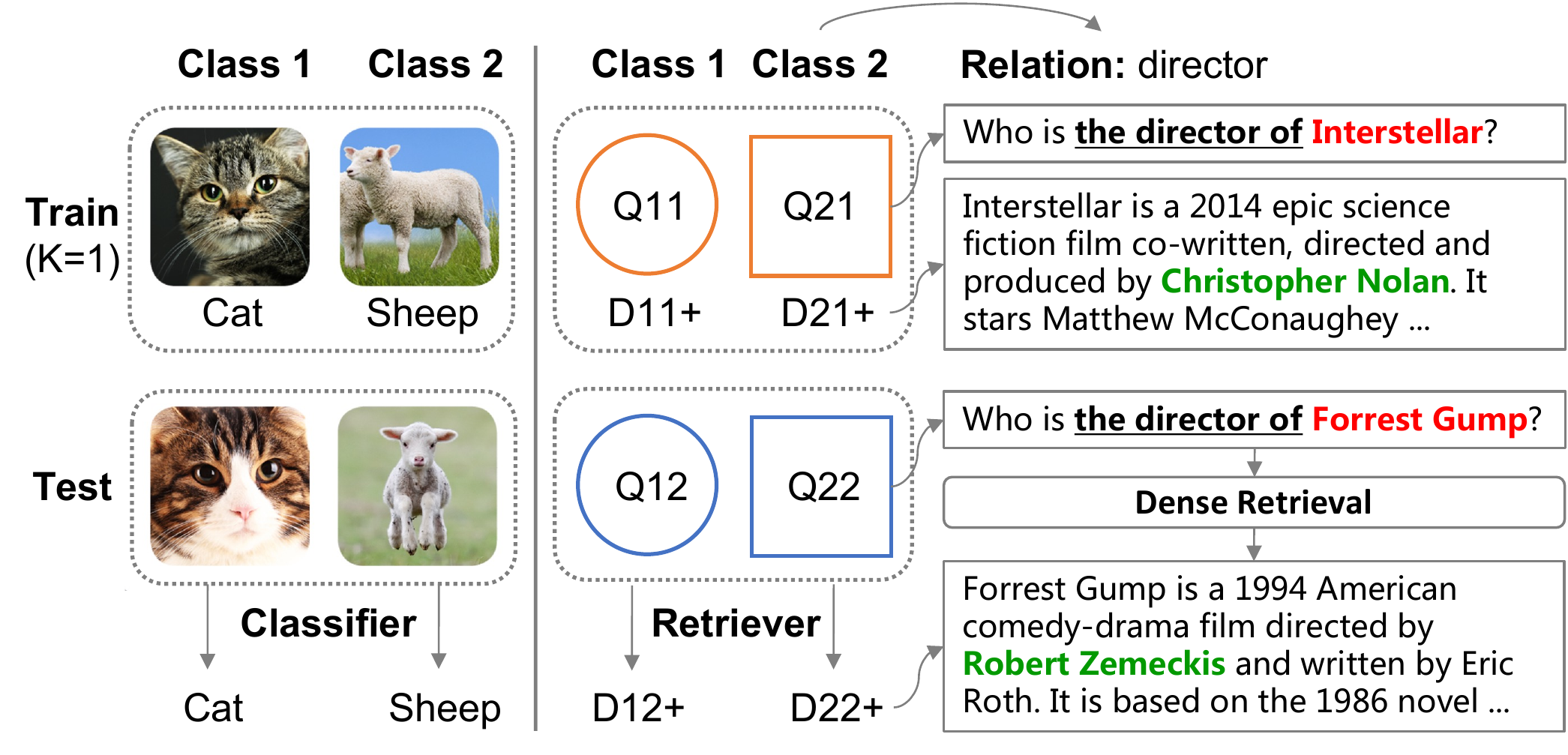}
\caption{\label{fig:framework}Few-shot learning scenes. E.g., one-shot learning accesses only a single training sample per class (K=1). Left: image classification. Right: dense retrieval. We define the relations between \textcolor{red}{query entity} and \textcolor{darkgreen}{answer entity} as \textit{classes}, and the samples with different entity pairs as \textit{shots} in FewDR.}
\end{figure}



Dense retrieval (DR) maps queries and documents into a continuous representation space, i.e., dense vectors, and assesses their similarity, allowing the efficient identification of relevant information from large corpora~\cite{huang2013learning, lee2019latent}. In recent years, DR has shown impressive performance in various applications, such as ad-hoc retrieval~\cite{ance, ar2}, conversational search~\cite{yu2021convdr}, and question answering~\cite{dpr}. Nevertheless, requiring large-scale training data for optimal performance~\cite{dpr,craswell2020overview}, DR remains challenging for real-life scenarios with limited supervision signals~\cite{thakur2beir}, e.g., long-tail queries~\cite{tail-query}, and medical searches~\cite{trec-covid}, etc. Therefore, research on zero- or few-shot DR has emerged and become an important issue of academic community~\cite{thakur2beir,contriever,gtr,cocodr,promptagator}.

Although both zero- and few-shot abilities can improve DR's generality and scalability, few-shot DR has made less progress~\cite{promptagator}, in part due to the lack of a well-defined standard benchmark. Specifically, standard few-shot learning~\cite{chen2019closer} often involves training a model with sufficient samples from base classes and a few samples from novel classes, and evaluating the model on unseen samples of novel classes, with a ``shot'' being a single sample per class, as shown in Figure~\ref{fig:framework}. The standard setup would separate the base and novel classes, ensuring training and testing data include the same classes but different shots~\cite{wang2018low,chen2019closer}. However, current few-shot DR studies lack such benchmarks and often transform a ``few-data'' setting by randomly sampling small training data from a supervised dataset~\cite{sun2021few,hu2022p3,promptagator}. Consequently, it brings the following challenges in providing a robust and reliable few-shot evaluation for DR models:

First, supervised datasets are short of fine-grained class partitioning, so vanilla class-free random sampling would bring an inconsistent discrepancy between training and testing data, causing significant variations in test results when sampling different small training batches~\cite{liu2022challenges}. Second, current evaluation protocols of few-shot DR lack credibility due to the testing leakage risk and overlooking the importance of multiple assessments. E.g., \citet{promptagator} utilize a query generator trained on diverse supervised datasets~\cite{flan} to improve DR’s accuracy on novel tasks, which risks novel data leakage. Additionally, existing methods do not adequately account for the instability of fine-tuning with small data~\cite{dodge2020fine,zhangrevisiting}, so as to require multiple trials of sampling for more accurate estimation~\cite{gao2021making}.

To address the above challenges, we customize a standard FewDR dataset and evaluation protocol for few-shot dense retrieval. The dataset is constructed on the Wikipedia corpus, and contains 41,420 samples with 60 fine-grained classes. In FewDR, the relations between query-answer entities are modeled as ``classes'', while samples that involve different entity pairs but with the same relations are grouped as ``shots''. Figure~\ref{fig:framework} shows an example of FewDR. For standardized benchmarking, FewDR is split into training and testing shots, ensuring consistency in class distribution. It is then disjointed into base and novel classes for a continuous learning task: learning with sufficient shots from base classes and few-shot learning novel classes. Lastly, few-shot performance is estimated through multiple sampling trials, which reduces fluctuation in testing results and prevents novel data leakage for a robust and reliable benchmark. 

We conduct a systematic evaluation of DR models using the FewDR benchmark and discover that they fall short of their data-sufficient performance. We also thoroughly analyze these results in the hope of providing more insights for future work.

\section{FewDR Dataset}

\begin{table}[t]
\centering
\caption{\label{tab:dataset}The statistics of the FewDR dataset and the R@10 score of BM25 on the \#Test sets.}
\resizebox{\columnwidth}{!}{
\begin{tabular}{l c c c c c c}
\toprule
\textbf{Split} & \textbf{Classes} & \textbf{Queries} & \textbf{\#Train} & \textbf{\#Test} & \textbf{Corpus Size} & \textbf{Test R@10} \\
\hline
All & 60 & 41,420 & 20,726 & 20,694 & 21,015,324 & 75.9 \\
Base & 30 & 20,668 & 10,341 & 10,327 & 21,015,324 & 75.1 \\
Novel & 30 & 20,752 & 10,385 & 10,367 & 21,015,324 & 76.7 \\
\bottomrule
  \end{tabular}
} 
\end{table}




This section briefly introduces the process of creating the FewDR dataset. More construction details and statistics will be placed in our open-source repository due to space constraints.

\textbf{Dataset Construction}. The entire Wikipedia is our corpus and Wikidata~\cite{wikidata} is our aligned knowledge base with fact triples \textit{(head entity, relation, tail entity)}. FewDR is constructed in four steps: \textit{(1) Class Selection}: we first select 60 relation classes from Wikidata that are suitable as query intents. \textit{(2) Shot Collection}: Then we collect \textit{head/tail-entity} pairs as shots for these relations, removing duplicates, and equalizing the shots per class. \textit{(3) Query Synthesis}: queries are created by combining head entities and manual rewriting of question templates to mimic real-world queries, e.g., the relation ``director'' is rewritten as the template ``who is the director of \textit{head entity} ?''. \textit{(4) Passage Alignment}: We align fact triples to Wikipedia passages, ensuring that each query has at least one relevant passage.



\textbf{Dataset Splitation}. We randomly divide FewDR's 60 classes into 30 base classes and 30 novel classes with equal training and testing shots. Table~\ref{tab:dataset} shows the statistics. We find that BM25 performs similarly on base and novel classes, and particularly well on such entity-centric queries, which is consistent with previous work~\cite{entityquestion}.

\section{FewDR Benchmark}


This section first recaps the preliminary of DR. Next, we introduce the FewDR benchmark and conduct a robustness analysis on it.

\subsection{Preliminary of DR} 

The first stage of retrieval task aims to find a set of relevant documents $D^+$ from a corpus $C$, based on a given query $q$. In DR, the query $q$ and the document $d$ are encoded as dense vectors and the relevance score $f(q, d; \theta)$ is often computed by the dot-product ($\cdot$) similarity metric:
\begin{align}
    f(q, d; \theta) &= g(q; \theta) \cdot g(d; \theta), \label{eq.dualencoder}
\end{align}
where $g(\cdot;\theta)$ denotes the text encoder with parameters $\theta$, which is often initialized by pre-trained language models~\citep{lee2019latent}. 

The learning objective can be formulated as optimizing DR's parameters such that positive pairs $(q, d^{+})$ of the query and relevant documents have higher similarity than the negative ones $(q, d^{-})$:
\begin{small}
\begin{equation}
\begin{aligned}
\setlength{\abovedisplayskip}{1pt}
\setlength{\belowdisplayskip}{1pt}
\theta^{*} &= \arg \min _{\theta} \mathcal{L}(q, D^+, D^-; \theta),
\\
&= \arg \min _{\theta} \sum_{q; d^+ \in D^+} \sum_{d^- \in D^-} l(q, d^+, d^-; \theta),
\end{aligned}
\label{eq:contra} 
\end{equation}
\end{small}
\begin{small}
\begin{equation}
\begin{aligned}
\setlength{\abovedisplayskip}{1pt}
\setlength{\belowdisplayskip}{1pt}
l(q, d^+, d^-; \theta) = - \text{log}  \frac{\exp{(f(q, d^{+}; \theta))}}{\exp{(f(q, d^{+}; \theta))}+ \exp{(f(q, d^{-}; \theta)})},
\end{aligned}
\label{eq:contra-2}
\end{equation}
\end{small}
where negative documents $D^-$ are usually sampled from the rest corpus $C\setminus D^+$. Once trained, DR can be used to retrieve the Top N related documents for a given query: $\text{Top N}_{d \in C} f(q, d; \theta^{*})$.



\subsection{Benchmarking Few-Shot DR}
Our goal is to establish a benchmark for few-shot DR that mimics real-world scenarios. In many practical applications, DR models are trained on base query classes with large supervision and regularly encounter novel query classes from users. These applications require the ability to generalize to novel classes but often lack ample training data and retraining infrastructure. To mimic this scene, we benchmark few-shot DR as a two-stage continuous learning task:




\textbf{Basic Learning.} First, the DR model is trained with base classes $\mathcal{C}_\text{base}$ with sufficient samples per class $\mathcal{B}_\text{train} = \{q^{}_\mathcal{B}, D^+_\mathcal{B}, D^-_\mathcal{B}\}$. The basic learning objective is computed as follows:
\begin{small}
\begin{equation}
\begin{aligned}
\setlength{\abovedisplayskip}{1pt}
\setlength{\belowdisplayskip}{1pt}
\theta_\text{base} = \arg \min_{\theta} \mathcal{L}(\mathcal{B}_\text{train}; \theta).
\end{aligned}
\end{equation}
\end{small}


\textbf{Few-Shot Learning.} Following, the DR model continues to learn unseen novel classes $\mathcal{C}_\text{novel}$, where $\mathcal{C}_\text{base} \cap \mathcal{C}_\text{novel} = \emptyset$. Suppose there is a novel training dataset $\mathcal{N}_\text{train} = \{q^{}_\mathcal{N}, D^+_\mathcal{N}, D^-_\mathcal{N}\}$, and the DR model only has access to $K$ samples for each novel class $\widetilde{\mathcal{N}}_\text{train} = \{(q_i^{}, D_i^{+}, D_i^{-})\}^{|C_\text{novel}| * K}_{i=1}$, while it still can access to all base classes. In such a case, the DR model can continue to optimize through:
\begin{small}
\begin{equation}
\begin{aligned}
\setlength{\abovedisplayskip}{1pt}
\setlength{\belowdisplayskip}{1pt}
\theta^{*} = \arg \min_{\theta_\text{base}^{}} \mathcal{L}(\widetilde{\mathcal{N}}_\text{train} \cup \widetilde{\mathcal{B}}_\text{train} ; \theta_\text{base}^{}),
\end{aligned}
\end{equation}
\end{small}
where $\widetilde{\mathcal{N}}_\text{train}$/$\widetilde{\mathcal{B}}_\text{train}$ are class-wise sampled from $\mathcal{N}_\text{train}$/$\mathcal{B}_\text{train}$.


\textbf{Testing.} In testing, the DR model ($\theta^{*}$) is evaluated on the combined classes $\mathcal{C}_\text{base} \cup \mathcal{C}_\text{novel}$ with unseen testing samples $\mathcal{B}_\text{test} \cup \mathcal{N}_\text{test}$. To better measure the few-shot ability, we repeat the few-shot learning and testing stages for multiple trials ($n>1$), each time with a different sampling of $\widetilde{\mathcal{N}}_\text{train}^{j}$/$\widetilde{\mathcal{B}}_\text{train}^{j}$, where $j=\{1, ..., n\}$.

\begin{figure}[t]
\centering
\subcaptionbox{DPR (BERT)}{
    \centering
    \includegraphics[height=3.3cm]{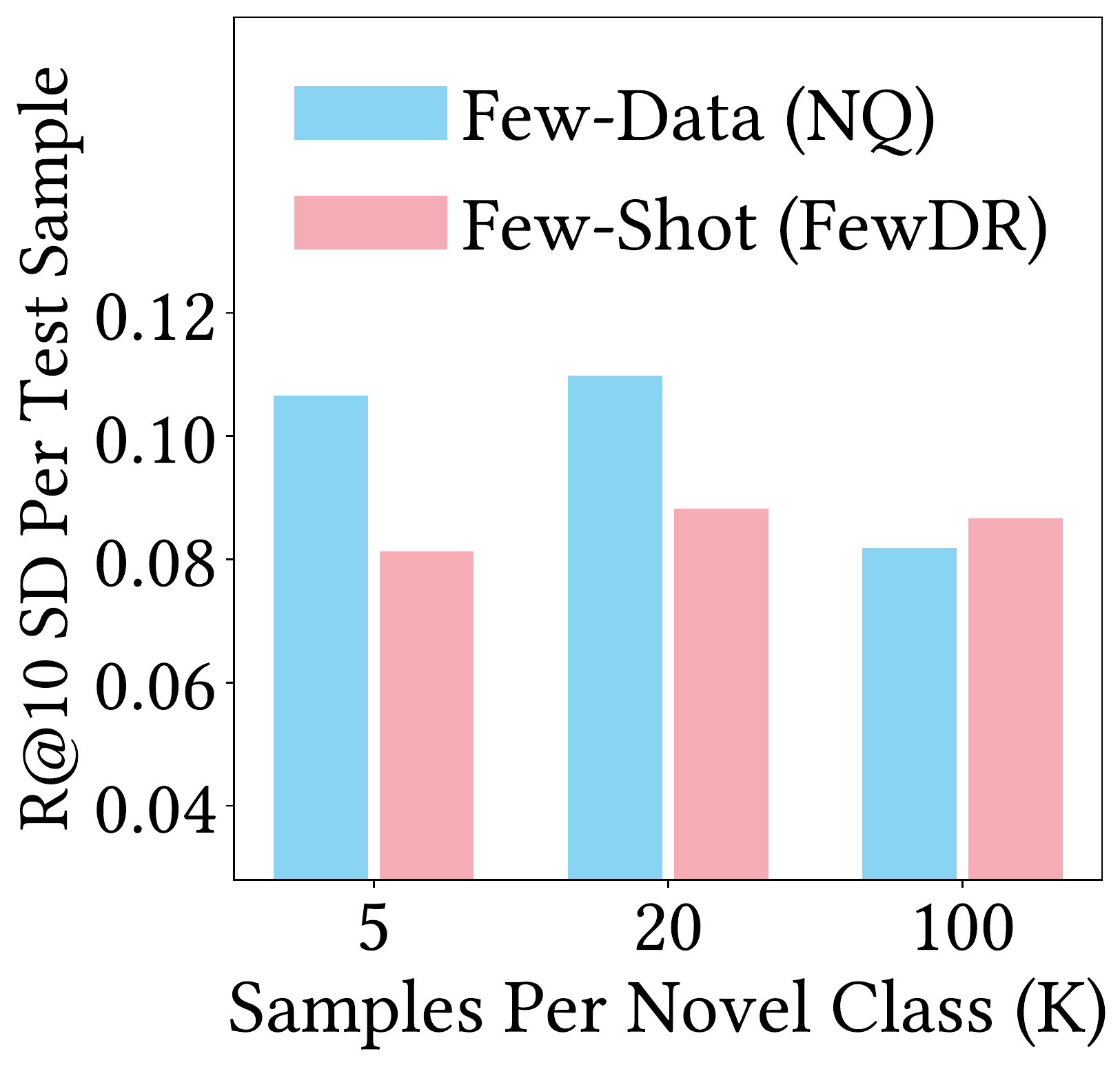}
}\hspace{0.15cm}
\subcaptionbox{DPR (coCondenser)}{
    \centering
    \includegraphics[height=3.3cm]{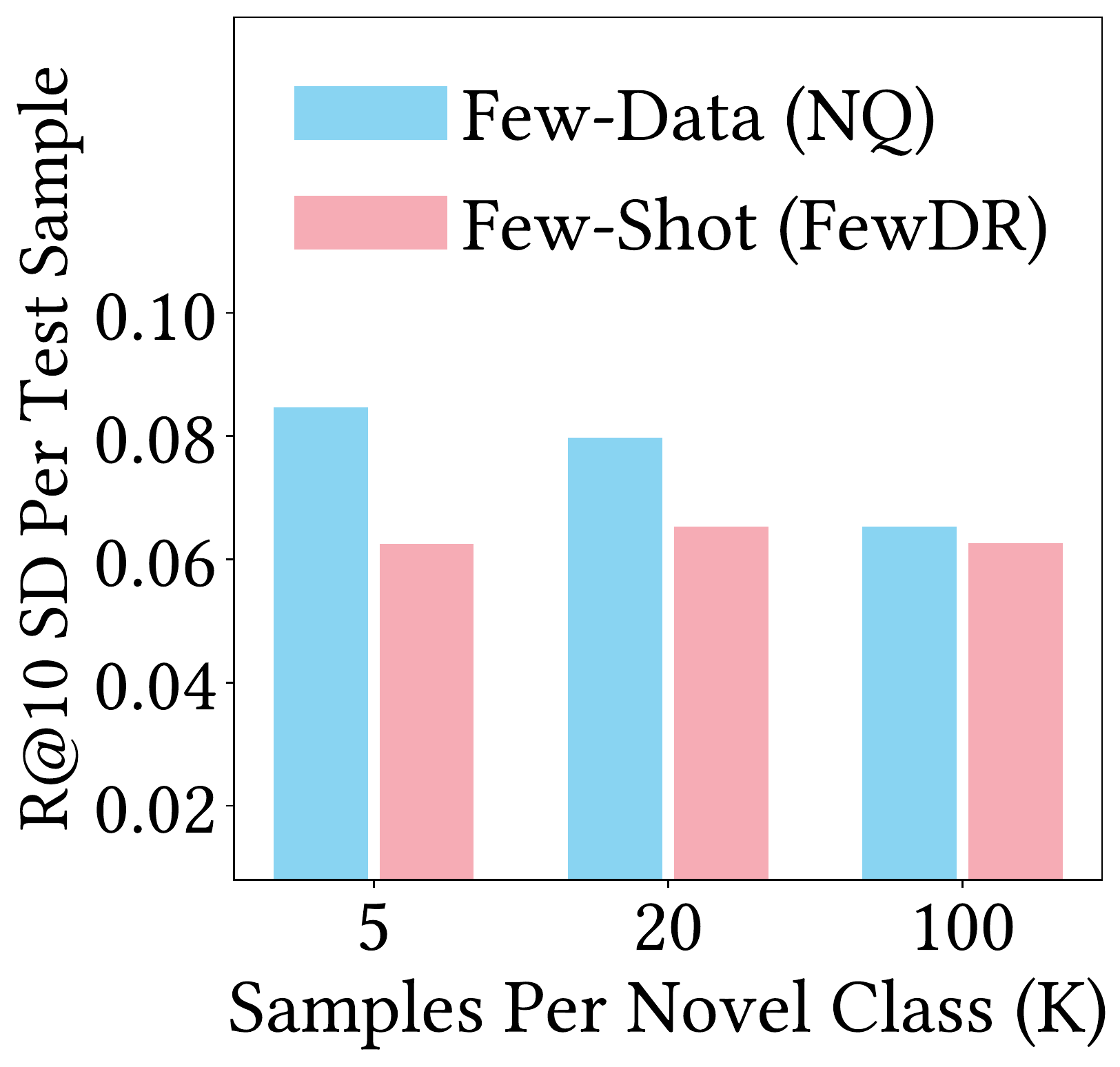}
}
\caption{\label{fig:robust}The standard deviation (SD) of the R@10 score on each test sample under five few-shot experimental trials.}
\end{figure}

\begin{table*}[t]
\centering
\vspace{-0.1cm}
\caption{\label{tab:overall}Few-shot DR results. We report the mean R@10 and standard deviation of five trials, on novel classes and all classes, for different few-shot degrees ($K$ means the number of samples per novel class). For a clear comparison, the results are grouped into three sheets, i.e., (1st) pre-training methods, (2nd) data augmentation and basic learning methods, and (3rd) few-shot learning methods. Superscripts indicate statistically significant improvements over (a)$\text{}^{a}$, (b)$\text{}^{b}$, (c)$\text{}^{c}$, (d)$\text{}^{d}$, (e)$\text{}^{e}$, (f)$\text{}^{f}$, (g)$\text{}^{g}$, (h)$\text{}^{h}$.} 
\vspace{-0.1cm}
\resizebox{\textwidth}{!}{
\begin{tabular}{l l l l l|l l l l l|l l l l l}
\toprule
\multicolumn{2}{l}{\multirow{2}{*}{\textbf{STG$_\text{pretrain}$}}} 
& \multirow{2}{*}{\textbf{STG$_\text{basic}$}} 
& \multirow{2}{*}{\textbf{STG$_\text{fewshot}$}}
& \multirow{2}{*}{\textbf{Augment}}
& \multicolumn{5}{c|}{\textbf{Novel}} 
& \multicolumn{5}{c}{\textbf{All (Base $+$ Novel)}} 
\\
& & & & & $K$ = 0 & \makecell[l]{5} & \makecell[l]{20} & \makecell[l]{100} & \makecell[l]{full} & $K$ = 0 & \makecell[l]{5} & \makecell[l]{20} & \makecell[l]{100} & \makecell[l]{full}
\\
\hline
\multicolumn{10}{l}{\textbf{1st sheet: } different pre-trained models and pre-training methods}
\\
\hline
\textit{(a)} & BERT$_\text{base}$ & DPR & Full-Tune & -- &  62.0$^{c d}$ & 65.2$_{1.2}$$^{c d}$ & 68.2$_{0.7}$$^{c d}$ & 73.8$_{1.1}$$^{c d}$ & 77.5$^{c d}$ & 67.0$^{c d}$ & 68.7$_{1.1}$$^{c d}$ & 70.2$_{0.5}$$^{c d}$ & 73.5$_{0.9}$$^{c d}$ & 75.7$^{c d}$
\\
\textit{(b)} & BERT$_\text{large}$ & DPR & Full-Tune & -- & 63.6$^{a c d}$ & 66.4$_{1.3}$$^{a c d}$ & 69.3$_{0.9}$$^{a c d}$ & 74.9$_{0.7}$$^{a c d}$ & 79.5$^{a c d}$ & 68.1$^{a c d}$ & 69.6$_{1.1}$$^{a c d}$ & 70.9$_{0.9}$$^{a c d}$ & 74.2$_{0.6}$$^{a c d}$ & 77.6$^{a c d}$
\\
\textit{(c)} & T5$_\text{base}$ & DPR & Full-Tune & Multi-Dataset & 49.5$^{d}$ & 57.1$_{1.3}$$^{d}$ & 61.1$_{0.9}$$^{d}$ & 66.4$_{0.5}$$^{d}$ & 71.2$^{d}$ & 57.8$^{d}$ & 61.4$_{1.0}$$^{d}$ & 63.2$_{0.7}$$^{d}$ & 66.2$_{0.3}$$^{d}$ & 69.3$^{d}$
\\
\textit{(d)} & T5-v1.1$_\text{base}$ & DPR & Full-Tune & -- & 43.9 & 50.2$_{1.8}$ & 56.8$_{0.9}$ & 62.2$_{0.4}$ & 63.3 & 53.2 & 54.9$_{1.4}$ & 58.7$_{0.7}$ & 61.8$_{0.4}$ & 61.3
\\
\textit{(e)} & Cropping & DPR & Full-Tune & -- & 68.6$^{a b c d}$ & 70.8$_{1.1}$$^{a b c d}$ & 74.1$_{0.5}$$^{a b c d}$ & 78.2$_{0.4}$$^{a b c d}$ & 80.9$^{a b c d}$ & 72.5$^{a b c d}$ & 73.6$_{0.8}$$^{a b c d}$ & 75.3$_{0.4}$$^{a b c d}$ & 77.5$_{0.2}$$^{a b c d}$ & 79.2$^{a b c d}$
\\
\textit{(f)} & Condenser & DPR & Full-Tune & -- & 70.4$^{a b c d e}$ & 73.8$_{1.0}$$^{a b c d e}$ & 75.9$_{0.4}$$^{a b c d e}$ & 80.0$_{0.3}$$^{a b c d e}$ & 82.9$^{a b c d e}$ & 74.6$^{a b c d e}$ & 76.4$_{0.9}$$^{a b c d e}$ & 77.3$_{0.3}$$^{a b c d e}$ & 79.5$_{0.3}$$^{a b c d e}$ & 81.1$^{a b c d e}$
\\
\textit{(g)} & coCondenser & DPR & Full-Tune & -- & \textbf{73.8}$^{a b c d e f}$ & \textbf{75.9$_{1.0}$}$^{a b c d e f}$ & \textbf{78.7$_{0.6}$}$^{a b c d e f}$ & \textbf{82.1$_{0.4}$}$^{a b c d e f}$ & \textbf{84.7}$^{a b c d e f}$ & \textbf{77.2}$^{a b c d e f}$ & \textbf{78.2$_{0.9}$}$^{a b c d e f}$ & \textbf{79.6$_{0.6}$}$^{a b c d e f}$ & \textbf{81.5$_{0.2}$}$^{a b c d e f}$ & \textbf{83.0}$^{a b c d e f}$
\\
\hline
\multicolumn{10}{l}{\textbf{2nd sheet: } different data augmentation and basic learning methods}
\\
\hline
\textit{(a)} & BERT$_\text{base}$ & DPR & Full-Tune & -- &  62.0 & 65.2$_{1.2}$ & 68.2$_{0.7}$ & 73.8$_{1.1}$ & 77.5 & 67.0 & 68.7$_{1.1}$ & 70.2$_{0.5}$ & 73.5$_{0.9}$ & 75.7
\\
\textit{(b)} & BERT$_\text{base}$ & DPR & Full-Tune & MARCO & 67.5$^{a}$ & 69.5$_{1.6}$$^{a}$ & 72.5$_{0.9}$$^{a}$ & 77.1$_{0.7}$$^{a}$ & 80.1$^{a}$ & 71.7$^{a}$ & 72.7$_{1.3}$$^{a}$ & 74.1$_{0.6}$$^{a}$ & 76.5$_{0.6}$$^{a}$ & 78.5$^{a}$
\\
\textit{(c)} & BERT$_\text{base}$ & DPR & Full-Tune & QG (MARCO) & 77.4$^{a b e}$ & 78.5$_{0.7}$$^{a b e}$ & 80.3$_{0.4}$$^{a b e}$ & 83.1$_{0.4}$$^{a b e}$ & 85.9$^{a b e}$ & 80.0$^{a b e}$ & 80.3$_{0.6}$$^{a b e}$ & 81.1$_{0.3}$$^{a b e}$ & 82.5$_{0.3}$$^{a b e}$ & 84.2$^{a b e}$
\\
\textit{(d)} & BERT$_\text{base}$ & DPR & Full-Tune & MARCO+QG & 78.5$^{a b c e}$ & 79.5$_{0.5}$$^{a b c e}$ & 81.3$_{0.3}$$^{a b c e}$ & 83.8$_{0.3}$$^{a b c e}$ & 86.3$^{a b c e}$ & 80.8$^{a b c e}$ & 81.2$_{0.6}$$^{a b c e}$ & 82.0$_{0.2}$$^{a b c e}$ & 83.2$_{0.2}$$^{a b c e}$ & 84.7$^{a b c e}$
\\
\textit{(e)} & coCondenser & DPR & Full-Tune & -- & 73.8$^{a b}$ & 75.9$_{1.0}$$^{a b}$ & 78.7$_{0.6}$$^{a b}$ & 82.1$_{0.4}$$^{a b}$ & 84.7$^{a b}$ & 77.2$^{a b}$ & 78.2$_{0.9}$$^{a b}$ & 79.6$_{0.6}$$^{a b}$ & 81.5$_{0.2}$$^{a b}$ & 83.0$^{a b}$
\\
\textit{(f)} & coCondenser & ANCE & Full-Tune & -- & 78.1$^{a b e}$ & 81.0$_{0.5}$$^{a b c d e g}$ & 82.9$_{0.3}$$^{a b c d e}$ & 85.0$_{0.3}$$^{a b c d e}$ & \textbf{88.9}$^{a b c d e g}$ & 81.0$^{a b e}$ & \textbf{82.5$_{0.3}$}$^{a b c d e}$ & \textbf{83.5$_{0.2}$}$^{a b c d e h}$ & 84.6$_{0.1}$$^{a b c d e}$ & \textbf{87.3}$^{a b c d e g}$
\\
\textit{(g)} & coCondenser & ANCE-Tele & Full-Tune & -- & 78.5$^{a b e}$ & 80.3$_{0.1}$$^{a b c e}$ & 82.4$_{0.3}$$^{a b c d e}$ & \textbf{85.4$_{0.3}$}$^{a b c d e}$ & 87.9$^{a b c d e}$ & 81.4$^{a b c e f}$ & 82.4$_{0.1}$$^{a b c d e}$ & \textbf{83.5$_{0.2}$}$^{a b c d e}$ & \textbf{85.1$_{0.2}$}$^{a b c d e f h}$ & 86.3$^{a b c d e}$
\\
\textit{(h)} & coCondenser & Distil-ANCE & Full-Tune & -- & \textbf{80.4}$^{a b c e f g}$ & \textbf{81.4$_{0.3}$}$^{a b c d e g}$ & \textbf{83.2$_{0.2}$}$^{a b c d e}$ & 85.1$_{0.2}$$^{a b c d e}$ & 88.5$^{a b c d e}$ & \textbf{81.6}$^{a b c e}$ & 82.1$_{0.3}$$^{a b c d e}$ & 83.0$_{0.2}$$^{a b c d e}$ & 84.6$_{0.1}$$^{a b c d e}$ & 87.0$^{a b c d e g}$
\\ 
\hline
\multicolumn{10}{l}{\textbf{3rd sheet: } different fine-tuning methods in the few-shot learning stage} 
\\
\hline
\textit{(a)} & coCondenser & DPR & Full-Tune & -- & \textbf{73.8} & \textbf{75.9$_{1.0}$}$^{b c d e}$ & \textbf{78.7$_{0.6}$}$^{b c d e}$ & \textbf{82.1$_{0.4}$}$^{b c d e}$ & \textbf{84.7}$^{b d e}$ & \textbf{77.2} & \textbf{78.2$_{0.9}$}$^{b c d e}$ & \textbf{79.6$_{0.6}$}$^{b c d e}$ & \textbf{81.5$_{0.2}$}$^{b c d e}$ & \textbf{83.0}$^{b d e}$
\\
\textit{(b)} & coCondenser & DPR & Qry-Tune & -- & \textbf{73.8} & 72.5$_{0.9}$ & 75.0$_{0.2}$$^{d e}$ & 77.4$_{0.4}$$^{d}$ & 78.4 & \textbf{77.2} & 75.6$_{0.8}$ & 77.1$_{0.2}$ & 78.4$_{0.3}$$^{d}$ & 79.2
\\
\textit{(c)} & coCondenser & DPR & Psg-Tune & -- & \textbf{73.8} & 75.4$_{1.6}$$^{b d e}$ & 77.6$_{0.3}$$^{b d e}$ & 81.2$_{0.2}$$^{b d e}$ & 84.3$^{b d e}$ & \textbf{77.2} & 77.7$_{1.2}$$^{b d e}$ & 78.9$_{0.4}$$^{b d e}$ & 81.1$_{0.1}$$^{b d e}$ & 82.6$^{b d e}$
\\
\textit{(d)} & coCondenser & DPR & BitFit & -- & \textbf{73.8} & 73.6$_{0.2}$$^{b}$ & 73.5$_{0.2}$ & 74.9$_{0.3}$ & 77.3 & \textbf{77.2} & 77.0$_{0.1}$$^{b}$ & 76.8$_{0.1}$ & 77.5$_{0.2}$ & 79.2
\\
\textit{(e)} & coCondenser & DPR & Prefix & -- & \textbf{73.8} & 73.6$_{0.2}$$^{b}$ & 73.7$_{0.2}$ & 76.8$_{0.2}$$^{d}$ & 79.6$^{b d}$ & \textbf{77.2} & 77.0$_{0.2}$$^{b}$ & 77.0$_{0.2}$ & 78.6$_{0.2}$$^{d}$ & 80.2$^{b d}$
\\
\bottomrule
\end{tabular}
}
\end{table*}
\vspace{-0.2cm}






\subsection{Benchmark Investigation}

The FewDR benchmark, based on the FewDR dataset and the corresponding benchmarking protocol, is compared to the ``few-data'' NQ benchmark transformed from the NQ dataset~\cite{nq}. Stability is assessed through a set of few-shot experiments on DPR~\cite{dpr} with R@10 standard deviation (SD) calculated for each test sample across five trials and results compared across different pre-trained models. As Figure~\ref{fig:robust} depicts, our FewDR benchmark shows notably lower SD, especially in the 5/20-shot scenarios, indicating its effectiveness in robustly assessing DR's few-shot performance with fewer trials.

\section{FewDR Experiments}

In this section, we employ the FewDR benchmark to evaluate current SOTA DR methods. We first introduce these DR methods and implementation details, and then analyze the evaluation results. More experimental details will be outlined in our public repository.

\subsection{DR Models}

Recently, DR research has made strides in diverse directions, such as DR-oriented pre-training and data augmentation. We conduct experiments using a total of 17 DR models based on these directions. 



\textbf{Pre-training Methods.} We evaluate the few-shot performance of DPR~\cite{dpr} using various pre-trained models, including widely-used architectures (\texttt{BERT}~\cite{bert}, \texttt{T5}~\cite{t5}) with different sizes (\texttt{Base}, \texttt{Large}), and typical DR-oriented pre-training methods. These pre-training methods improve text representation by creating information bottlenecks (\texttt{Condenser}~\cite{condenser}) in the Transformer~\cite{attention} or by enhancing semantic text matching through contrastive learning (\texttt{Cropping}~\cite{contriever}). And \texttt{coCondenser}~\cite{cocondenser} combines both approaches.


\textbf{Data Augment Methods.} We evaluate various augmentation techniques in pre-training and basic learning stages. \texttt{T5}~\cite{t5} augments pre-training with multiple supervised datasets, while \texttt{T5-v1.1} excludes any supervised training. In basic learning, DR models often add heterogeneous datasets or query generators to create homogeneous data~\cite{thakur2beir}. We test such augmenting methods using the \texttt{MARCO} dataset~\cite{msmarco}, generated pseudo data (\texttt{QG}~\cite{thakur2beir}), and the hybrid version (\texttt{MARCO+QG}) where the query generator is trained on MARCO.


\textbf{Basic Learning Methods.} We compare DR models in different training negatives and query-document interactions. \texttt{DPR}~\cite{dpr} uses negatives mined from BM25~\cite{bm25}; \texttt{ANCE}~\cite{ance} leverages BM25 and self-mined negatives; \texttt{ANCE-Tele}~\cite{ancetele} utilizes self-mined negatives only. We also implement \texttt{Distil-ANCE}, which involves a re-ranker with a cross-encoder architecture for knowledge distillation in ANCE training based on previous work~\cite{rocketqav2}. 



\textbf{Few-Shot Learning Methods.} Our experiments also compare five fine-tuning methods in the few-shot learning stage. \texttt{Full-Tune} stands for full-parameter fine-tuning, while \texttt{Qry-Tune} and \texttt{Psg-Tune} tune only the query or passage encoder. In addition, recent studies~\cite{he2021effectiveness,delta} show that certain parameter-efficient tuning (PET) methods outperform vanilla fine-tuning in few-shot scenes. We thus take BitFit~\cite{bitfit} and Prefix~\cite{prefix} into account, representing specification-based and addition-based PET methods~\cite{delta}, respectively.

\begin{figure*}[t]
  \centering
  \hspace{-0.2cm}
    \begin{subfigure}[t]{0.19\linewidth}
        \includegraphics[height=3.1cm]{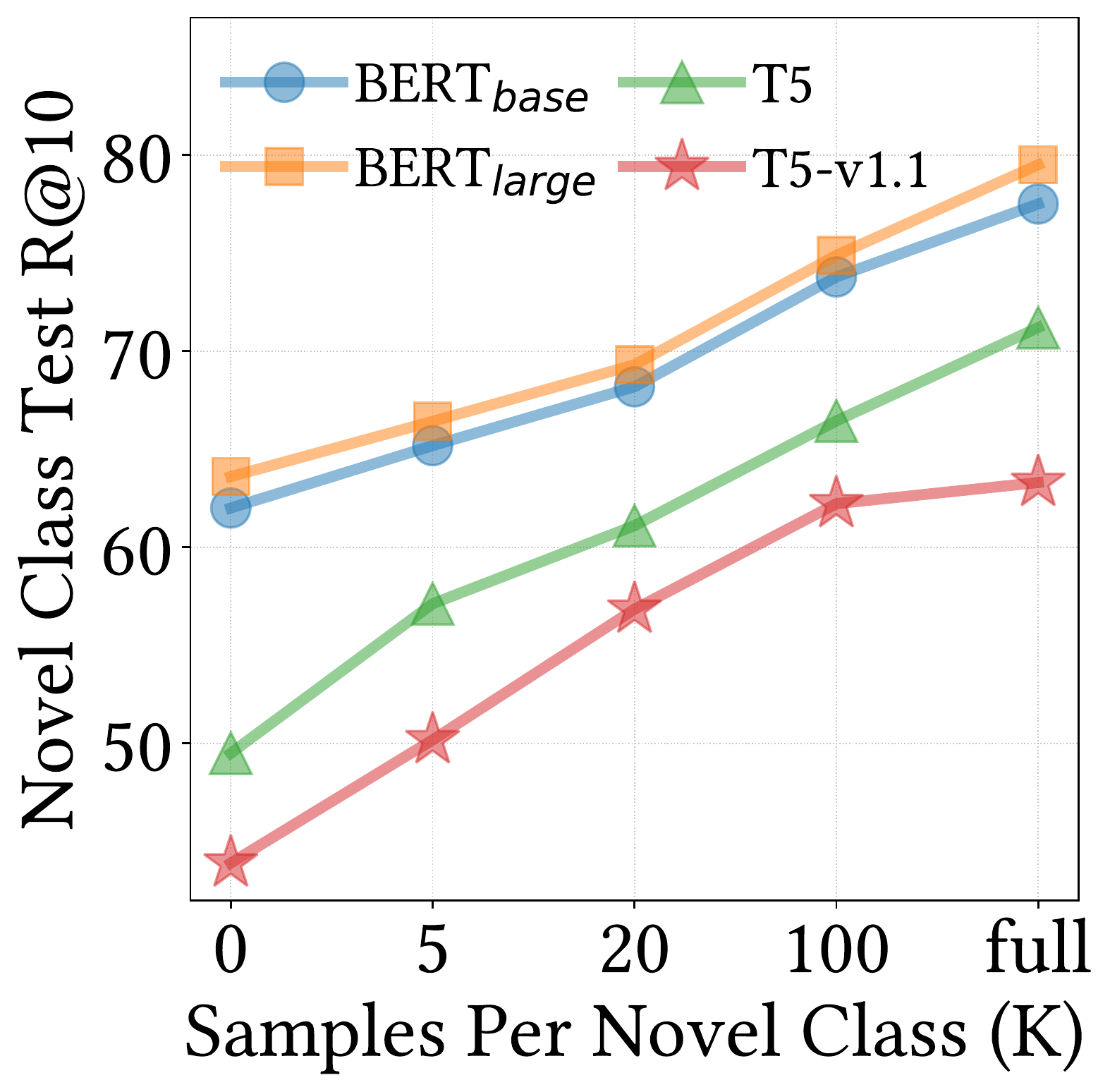}
        \caption{\label{subfig:pretrain-model-curve}Pre-trained Models.}
    \end{subfigure}
    \hspace{0.05cm}
    \begin{subfigure}[t]{0.19\linewidth}
        \includegraphics[height=3.1cm]{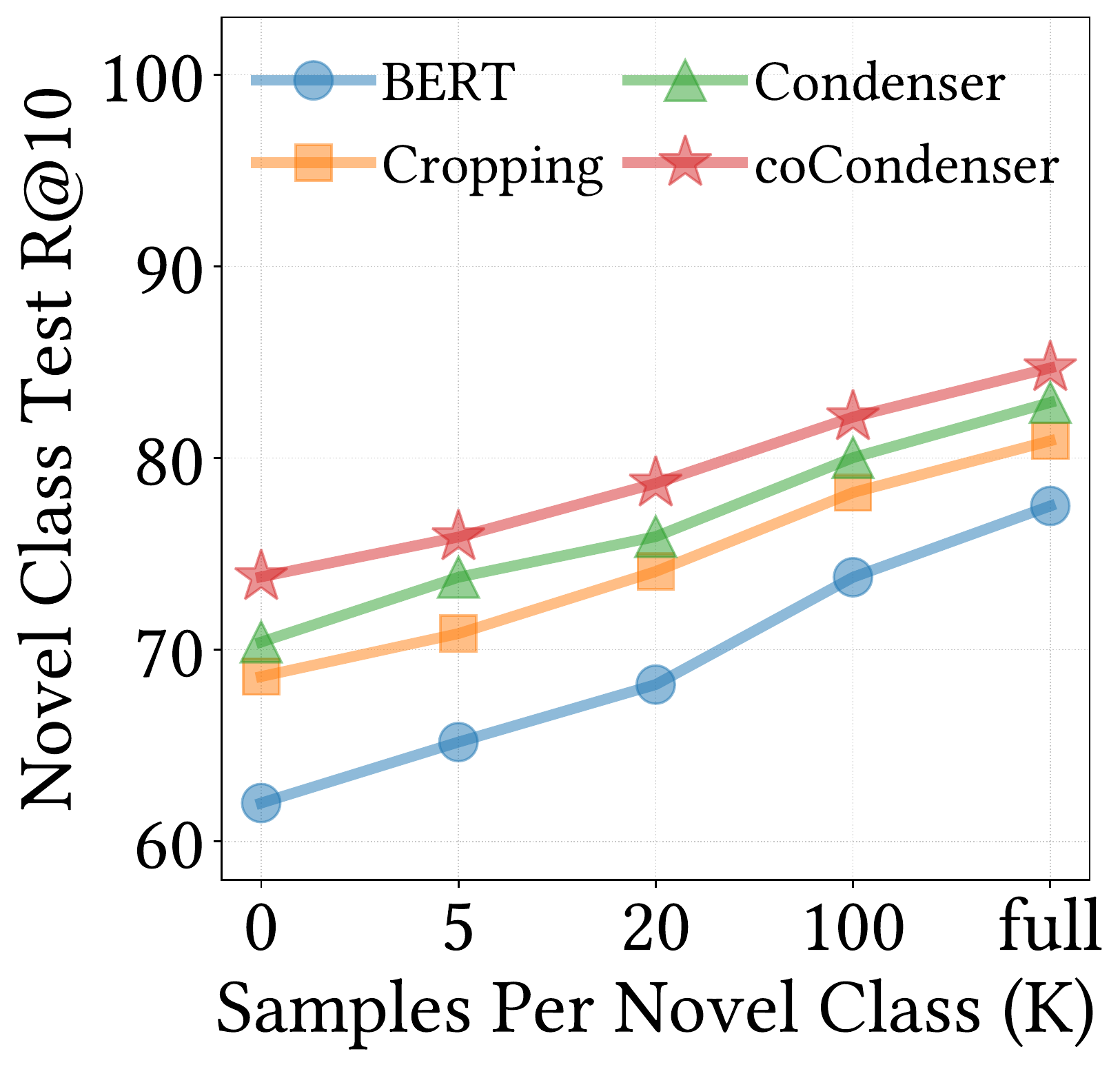}
        \caption{\label{subfig:pretrain-method-curve}Pre-training Ways.}
    \end{subfigure}
    \hspace{0.05cm}
    \begin{subfigure}[t]{0.19\linewidth}
        \includegraphics[height=3.1cm]{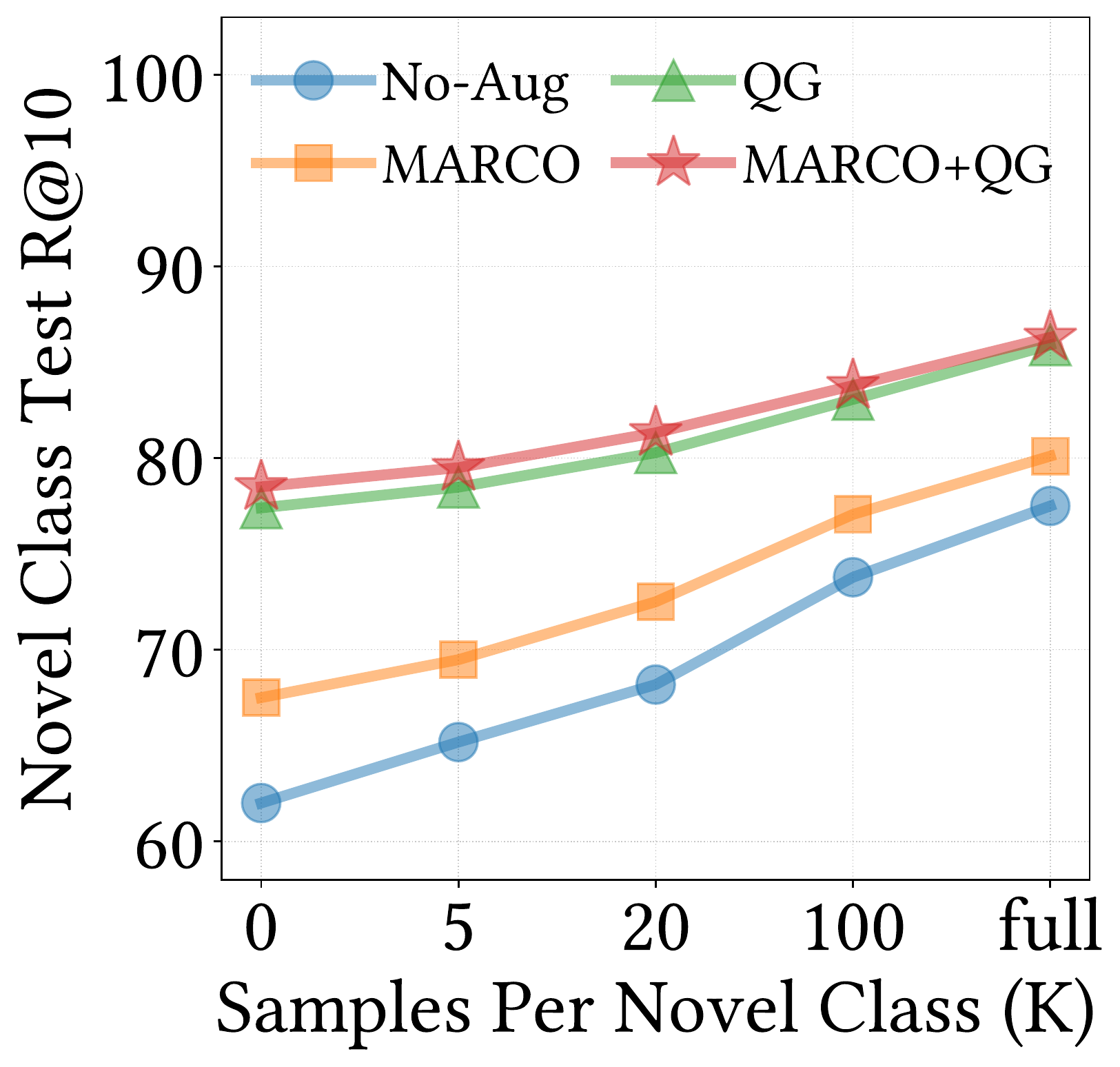}
        \caption{\label{subfig:data-augment-curve}Data Augmentation.}
    \end{subfigure}
    \hspace{0.05cm}
    \begin{subfigure}[t]{0.19\linewidth}
        \includegraphics[height=3.1cm]{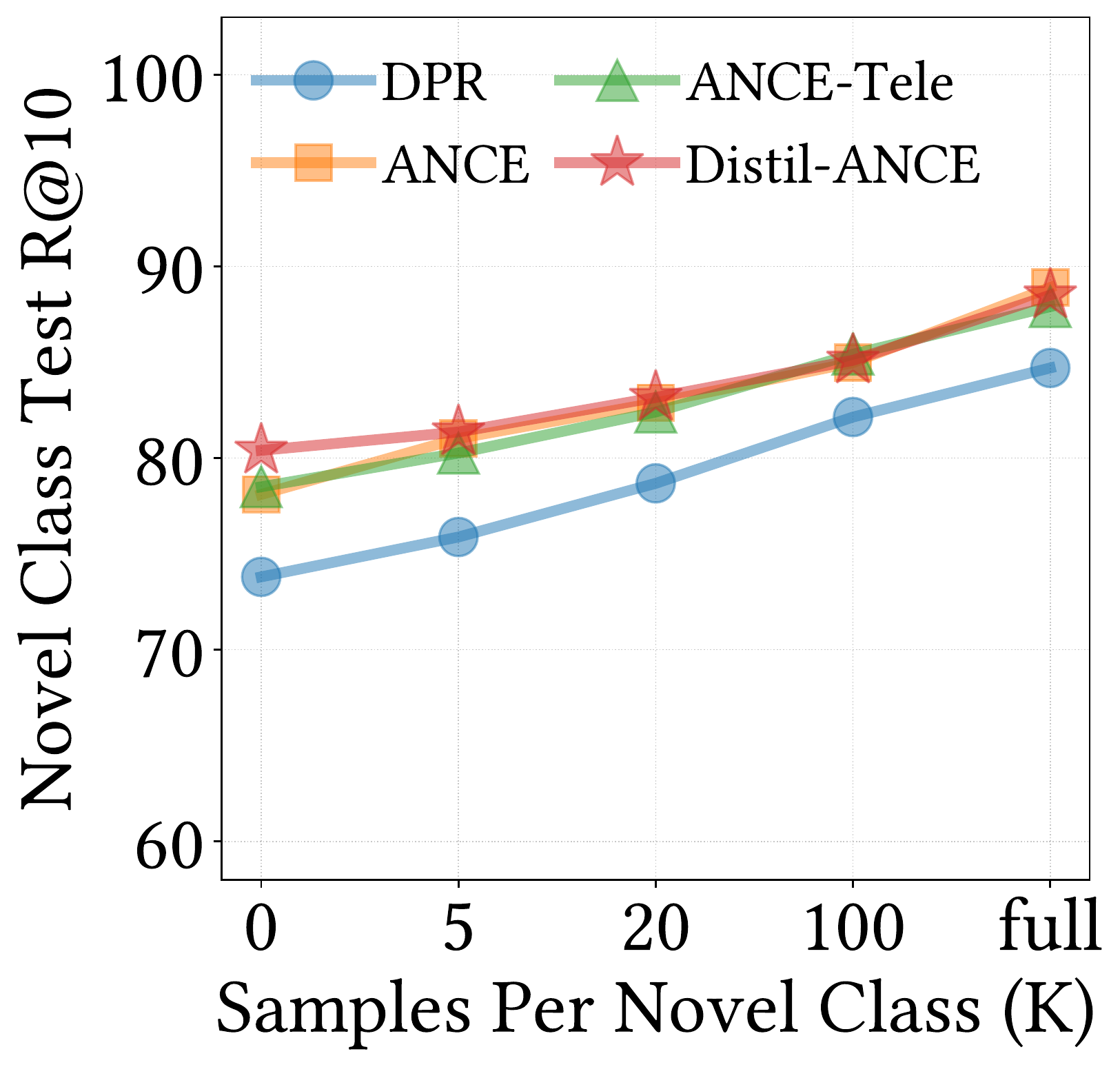}
        \centering
        \caption{\label{subfig:basic-curve}Basic Learning.}
    \end{subfigure}
    \hspace{0.05cm}
    \begin{subfigure}[t]{0.19\linewidth}
        \includegraphics[height=3.1cm]{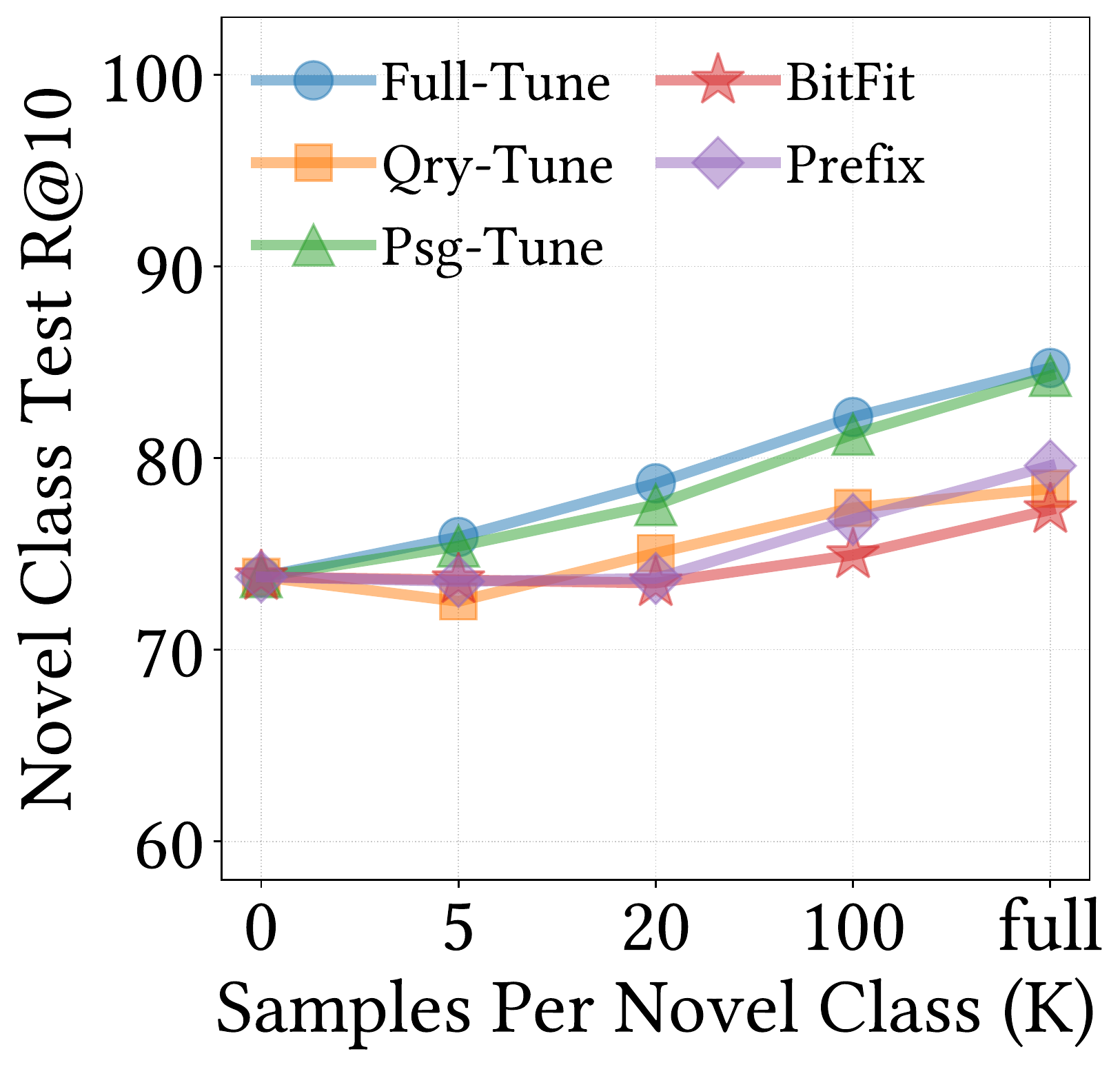}
        \centering
        \caption{\label{subfig:few-shot-curve}Few-Shot Learning.}
    \end{subfigure}
    \hspace{-0.4cm}
\caption{\label{fig:novel-ana} Few-shot boosting gradient on novel classes. X-axes denote few-shot degrees. Y-axes denote testing R@10 scores.}
\end{figure*}

\begin{figure*}[t]
  \centering
  \hspace{-0.2cm}
    \begin{subfigure}[t]{0.19\linewidth}
        \includegraphics[height=3.1cm]{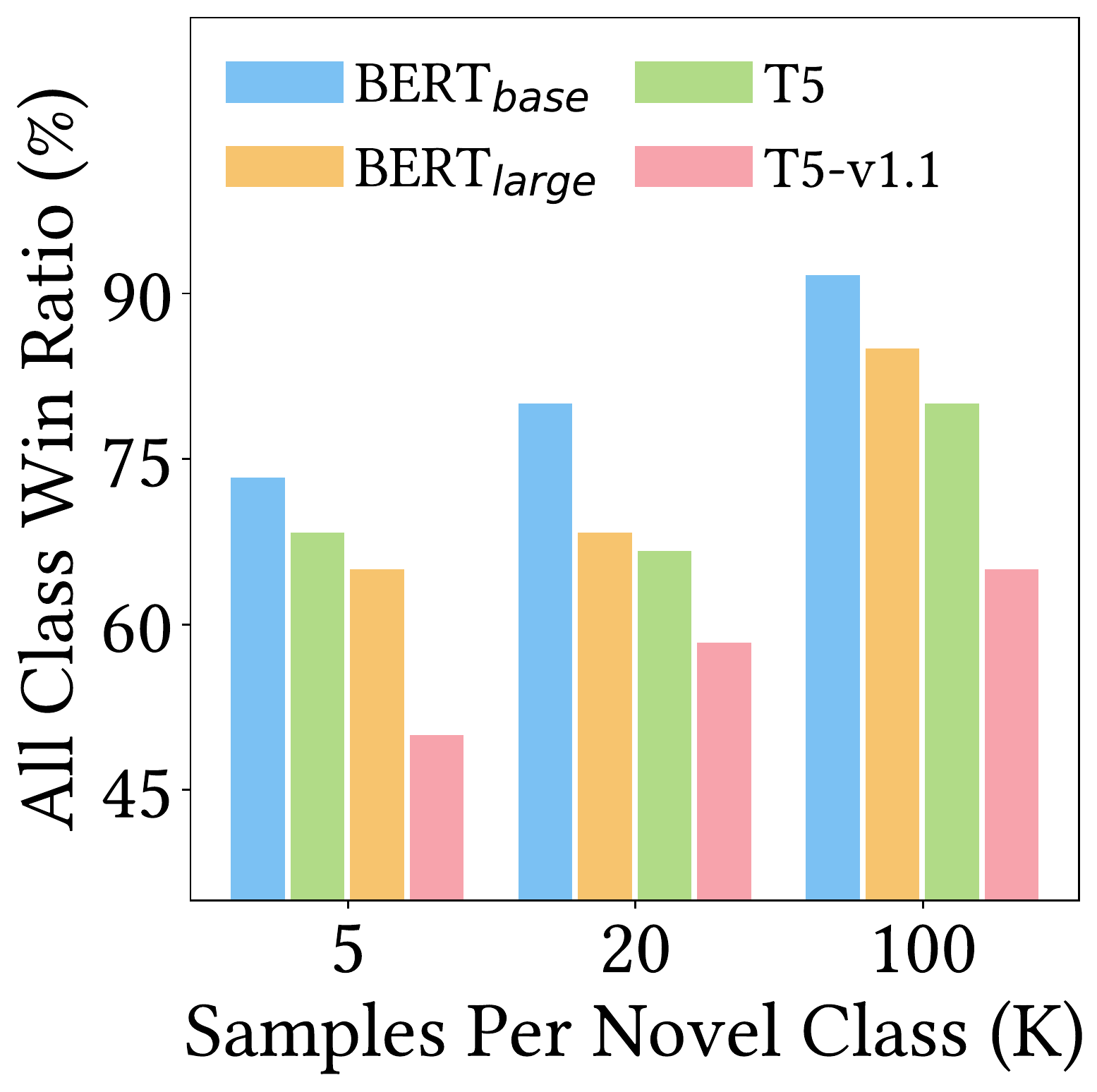}
        \caption{\label{subfig:pretrain-model-hist}Pre-trained models.}
    \end{subfigure}
    \hspace{0.05cm}
    \begin{subfigure}[t]{0.19\linewidth}
        \includegraphics[height=3.1cm]{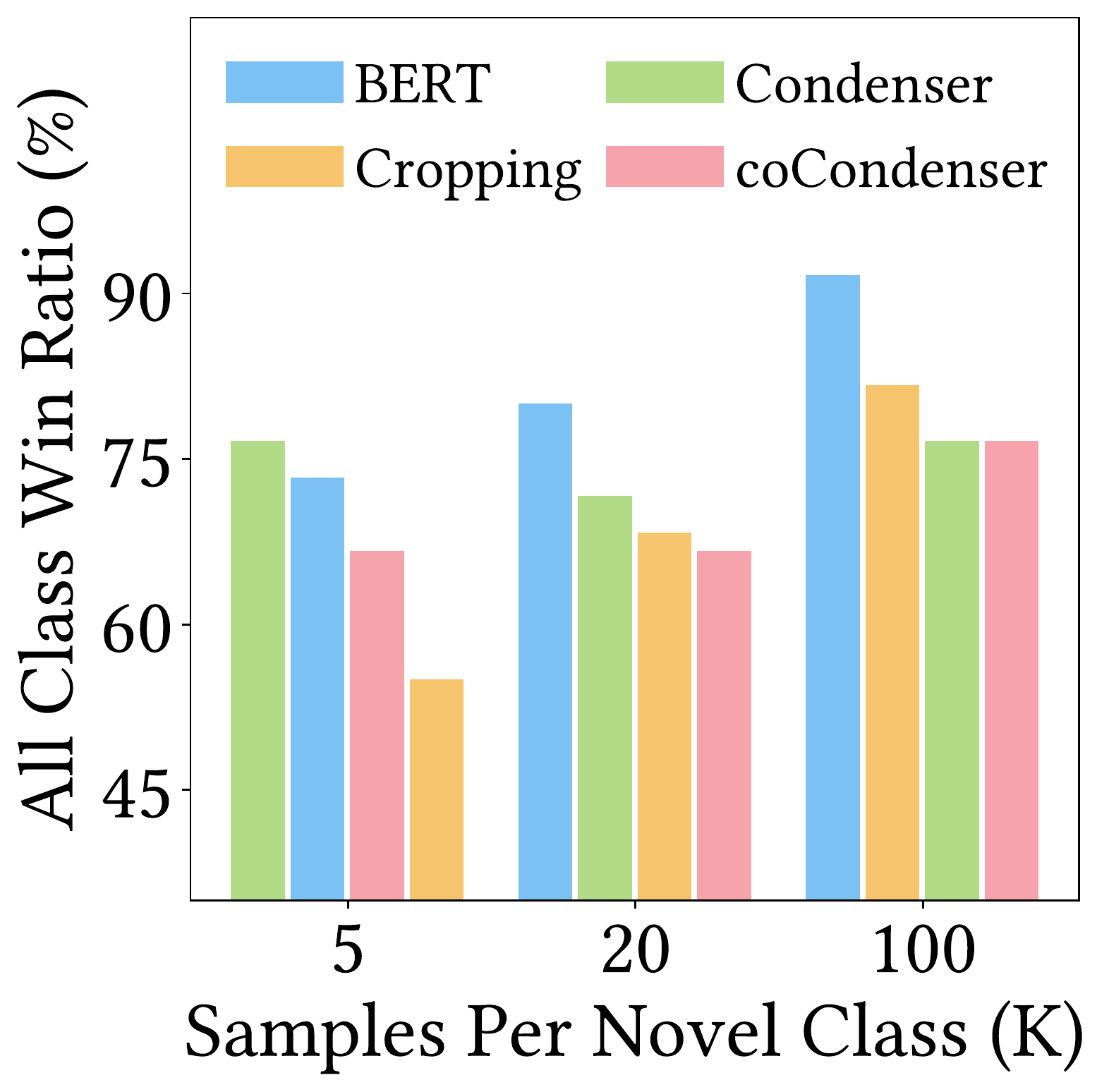}
        \caption{\label{subfig:pretrain-method-hist}Pre-training Ways.}
    \end{subfigure}
    \hspace{0.05cm}
    \begin{subfigure}[t]{0.19\linewidth}
        \includegraphics[height=3.1cm]{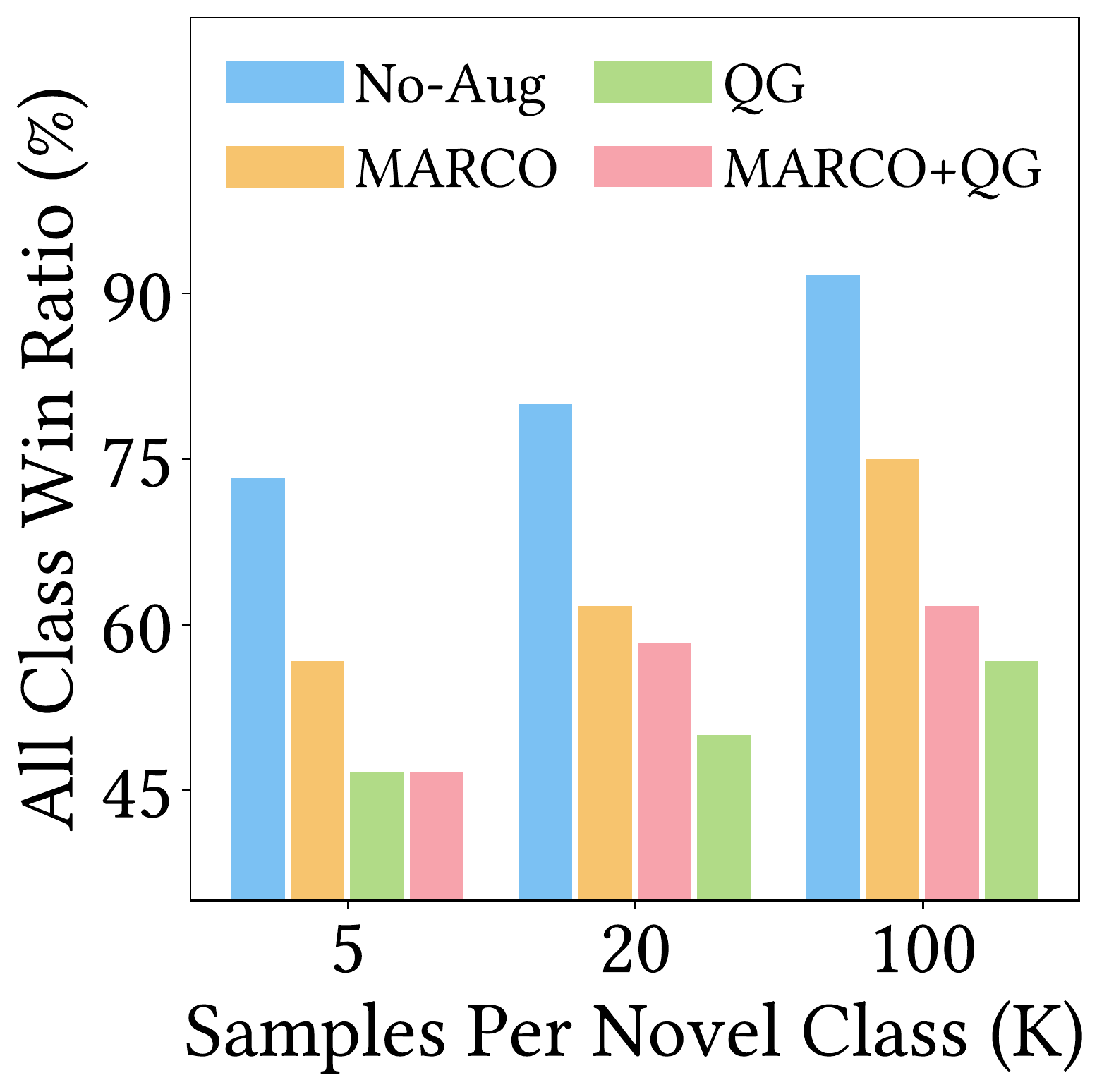}
        \caption{\label{subfig:augment-hist}Data Augmentation.}
    \end{subfigure}
    \hspace{0.05cm}
    \begin{subfigure}[t]{0.19\linewidth}
        \includegraphics[height=3.1cm]{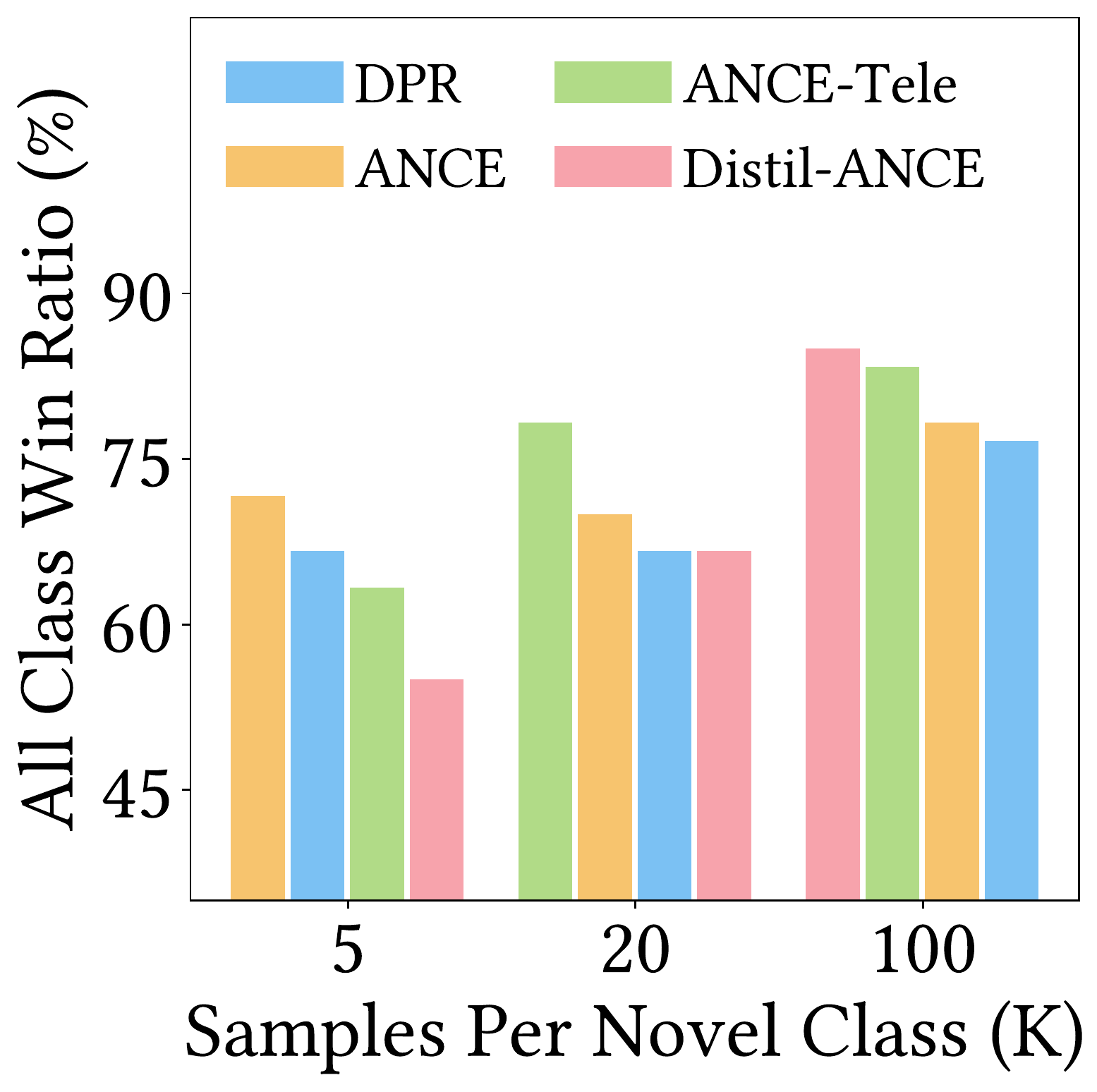}
        \centering
        \caption{\label{subfig:basic-hist}Basic Learning.}
    \end{subfigure}
    \hspace{0.05cm}
    \begin{subfigure}[t]{0.19\linewidth}
        \includegraphics[height=3.1cm]{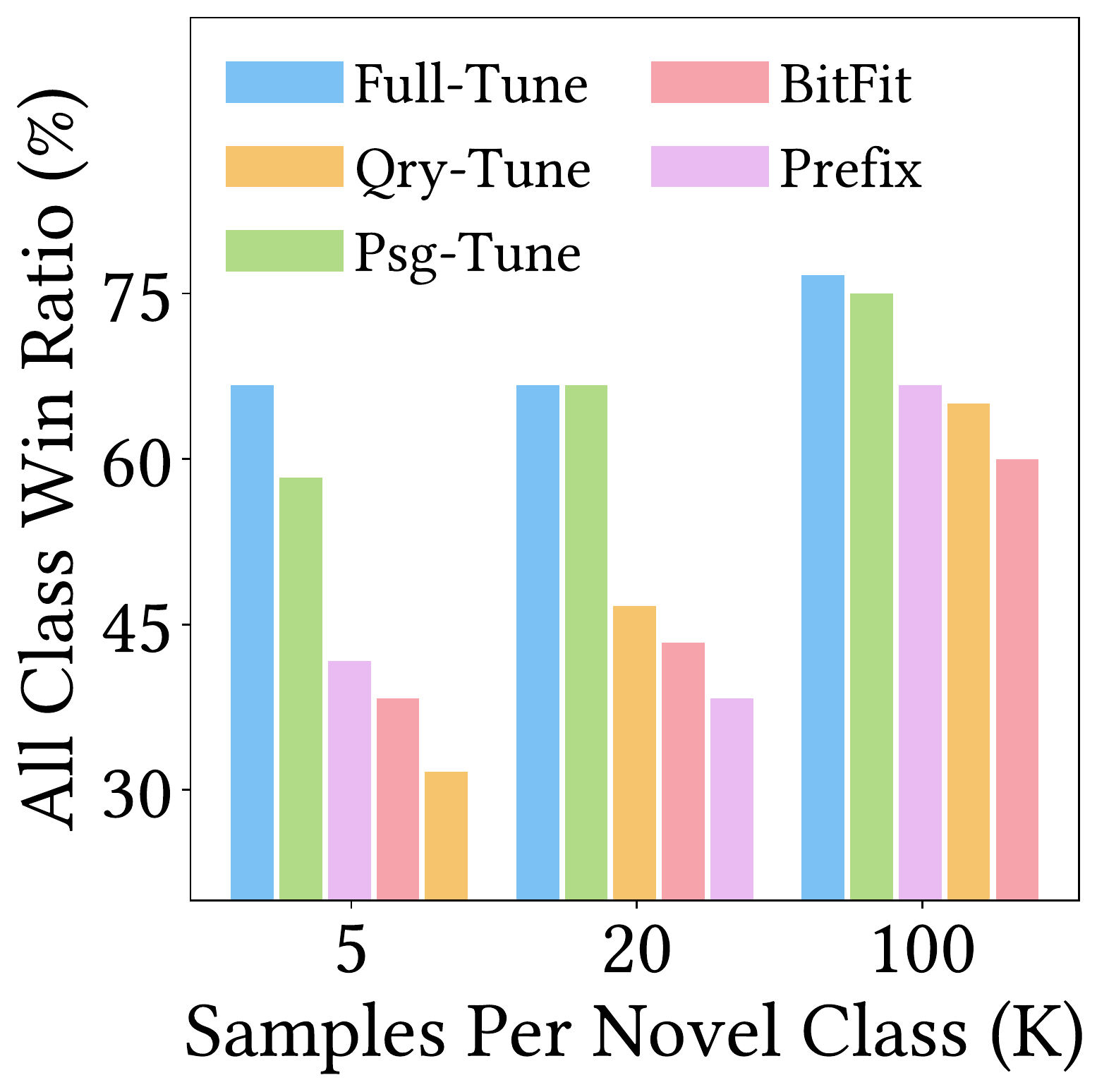}
        \centering
        \caption{\label{subfig:few-shot-hist}Few-Shot Learning.}
    \end{subfigure}
    \hspace{-0.4cm}
\caption{\label{fig:all-ana}Win ratio on all classes. X-axes denote few-shot degrees. Y-axes are the ratio of classes with higher R@10 scores (\textit{vs.} zero-shot performance).}
\end{figure*} 

\textbf{Implementation Details.} For all DR methods, we set the maximum lengths of query and passage to 32 and 256. During training, DPR undergoes a single training episode; \texttt{ANCE}, \texttt{ANCE-Tele}, and \texttt{Distil-ANCE} train iteratively with 3 episodes. Each episode involves fine-tuning the DR model for 40 epochs, with an AdamW optimizer, a batch size of 32, and a learning rate of 5e-6. We initialize the \texttt{Distil-ANCE}’s re-ranker with \texttt{BERT}$_\text{base}$, training it for 10 epochs with a learning rate of 5e-6. All training is done on a single A100 GPU (40G). We perform five trials per evaluation ($n=5$), reporting the mean and standard deviation (SD) of R@10 scores. Statistical significance is examined by permutation with $p<0.05$.

\subsection{FewDR Results}


\textbf{Overall Results.} Table~\ref{tab:overall} shows the overall results. On novel classes, few-shot performance for all DR models lags significantly behind full-shot, especially in 5-shot, with an average reduction of 23.1\%. The performance of nearly all models exhibits greater fluctuations in fewer shot scenarios, with the SD at 5-shot exceeding twice that of 100-shot. These results indicate the need for further research to enhance DR's generalization and robustness in few-shot scenarios.



\textbf{Pre-training Results.} Table~\ref{tab:overall}'s 1st sheet shows that the large version of BERT only brings marginal gains of 1.6\%/1.1\% (novel/all classes) over the base version, while T5 exhibits lower performance, particularly T5-v1.1. Promisingly, we observe superior performance for the three DR-oriented pre-training methods, with coCondenser achieving the best results with a 14.4\%/12.7\% (novel/all) improvement. These findings suggest exploring better ways to utilize larger models and T5 architectures for few-shot DR.



\textbf{Data Augment Results.} The efficacy of augmentation is demonstrated in the 2nd sheet, with QG's homogeneous augmentation delivering superior few-shot gains (16.9\%/14.9\% for novel/all classes). Despite its effectiveness, cogitation of test leakage risk is necessary, e.g., T5's implicit augmentation and QG's indirect augmentation.



\textbf{Basic Learning Results.} As the 2nd sheet shows, ANCE, ANCE-Tele, and Distil-ANCE enhance DPR's few-shot accuracy by around 5.2\%/4.6\% (novel/all), and notably lessen testing SD. Interestingly, each has distinct specialties: ANCE-Tele is highly robust with a few-shot SD of 0.17 on all classes, ANCE outperforms in full-shot, and Distil-ANCE recognizes novel classes better in zero-shot scenes.




\textbf{Few-Shot Learning Results.} Based on the 3rd sheet, full parameter tuning followed by tuning only the passage encoder produces the best outcomes, while query-only tuning lags far behind. These results align with previous study~\cite{entityquestion}, indicating that a robust corpus representation is critical for rapid adaptation to novel scenarios. However, in real-world applications, re-encoding all passages for adapting each batch of novel queries is inefficient, especially when dealing with vast corpora or larger models. There are also minimal gains for both PET methods in 5/20-shot scenes, necessitating a reassessment of PET methods designed for DR with larger models.





\textbf{Few-Shot Boosting Gradient.} Figure~\ref{fig:novel-ana} plots the few-shot accuracy curves of these DR models on novel classes. Despite varying few-shot accuracy, they show comparable efficiency in utilizing a small number of novel samples, i.e., with similar boosting gradients (slopes). Thereby, the few-shot accuracy of these models is largely determined by their zero-shot performance. It is also notable that the performance gap between the DR models shrinks as the sample size increases, suggesting that the advantage of massive supervision may smooth out the differences between various DR models.




\textbf{Few-Shot Winning Ratio.} Figure~\ref{fig:all-ana} presents the percentage of all classes that outperform their zero-shot accuracy. 
Most DR models exhibit lower ratio of winning classes in fewer shot scenarios, and none of them are capable of improving the accuracy of 100\% classes. E.g, Condenser (in Figure~\ref{subfig:pretrain-method-hist}) achieves the highest winning ratio in the 5-shot scenario, but it still falls short of exceeding 80\%. It's also worth noting that DR models vary winner rankings across different few-shot degrees, as shown in Figure~\ref{subfig:basic-hist}, ANCE and ANCE-Tele are winners in 5-shot and 20-shot scenes, respectively, while Distil-ANCE wins in 100-shot scenes. These results emphasize the importance of accurately defining \textit{classes} and \textit{shots} in the few-shot DR benchmark to prevent class-specific testing bias and provide quantitative few-shot estimations.

\section{Conclusion}

This paper proposes a specialized benchmark, namely FewDR, to estimate the few-shot ability of DR robustly and reliably. Our extensive experiments on FewDR provide the following insights: (1) Current DR methods underperform in accuracy and stability in few-shot scenes. (2) They heavily ride on their zero-shot accuracy and have similar abilities to exploit a few novel samples. (3) Challenges in tuning large models and updating corpus representations efficiently hinder their adaptability to real-world few-shot scenes. 

\section*{Limitations}
This work is currently in progress, and as such, there are still some limitations.
One limitation is that FewDR's queries are template-style questions, which restricts their authenticity and complexity. Our future work will be committed to upgrading the FewDR dataset with real-world questions and complex query intents. Additionally, FewDR is currently limited to in-domain few-shot scenes, but we have plans to broaden its scope to include cross-domain few-shot retrieval scenarios in the future.



\bibliography{citations}
\bibliographystyle{acl_natbib}

\appendix

\end{document}